\documentclass{article} 
\usepackage{iclr2025_conference,times}

\usepackage{amsmath,amsfonts,bm}

\def\eqref#1{equation~\ref{#1}}

\def\1{\bm{1}}

\DeclareMathAlphabet{\mathsfit}{\encodingdefault}{\sfdefault}{m}{sl}
\SetMathAlphabet{\mathsfit}{bold}{\encodingdefault}{\sfdefault}{bx}{n}

\usepackage{hyperref}
\usepackage{url}
\usepackage{graphicx}
\usepackage{booktabs, multirow} 
\usepackage{longtable}
\usepackage{makecell} 
\usepackage{todonotes}

\title{Assessing Robustness to Spurious Correlations in Post-Training Language Models}

\author{Julia Shuieh\thanks{Equal contribution.} , Prasann Singhal\footnotemark[1] , Apaar Shanker\footnotemark[1], John Heyer, George Pu, \& Samuel Denton \\
\makebox[\textwidth][c]{Scale AI} \\
\makebox[\textwidth][c]{San Francisco, CA 94103, USA} \\
\makebox[\textwidth][c]{\texttt{\{apaar.shanker,sam.denton\}@scale.com}}
}

\iclrfinalcopy 
\begin{document}

\maketitle

\begin{abstract}
Supervised and preference-based fine-tuning techniques have become popular for aligning large language models (LLMs) with user intent and correctness criteria. However, real-world training data often exhibits spurious correlations—arising from biases, dataset artifacts, or other “shortcut” features—that can compromise a model’s performance or generalization. In this paper, we systematically evaluate three post-training algorithms—Supervised Fine-Tuning (SFT), Direct Preference Optimization (DPO), and KTO (Kahneman-Tversky Optimization)—across a diverse set of synthetic tasks and spuriousness conditions. Our tasks span mathematical reasoning, constrained instruction-following, and document-grounded question answering. We vary the degree of spurious correlation (10\% vs. 90\%) and investigate two forms of artifacts: “Feature Ambiguity” and “Distributional Narrowness.” Our results show that the models often but not always degrade under higher spuriousness. The preference-based methods (DPO/KTO) can demonstrate relative robustness in mathematical reasoning tasks. By contrast, SFT maintains stronger performance in complex, context-intensive tasks. These findings highlight that no single post-training strategy universally outperforms in all scenarios; the best choice depends on the type of target task and the nature of spurious correlations.
\end{abstract}

\section{Introduction}

Post-training alignment of large language models (LLMs) has emerged as a critical step in ensuring safe, accurate, and helpful responses \citep{zhang2023instruction}. Commonly used techniques include Supervised Fine-Tuning (SFT) on curated demonstrations and instruction data \citep{wei2021finetuned}, and preference-based methods (e.g., Direct Preference Optimization (DPO) \citep{rafailov2023direct}, Kahneman-Tversky Optimization (KTO) \citep{ethayarajh2024kto}. Although these methods produce impressive performance on various benchmarks, real-world training data often contain noisy or spurious correlations—features that are correlated with correctness in the training set but are not causally related to the target task. When the model overfits to these spurious patterns, it may underperform on broader distributions and degrade in accuracy under slight distribution shifts.

In this paper, we systematically investigate how different post-training approaches handle data contaminated by spurious correlations. We create a suite of controlled synthetic training sets with varied tasks, spuriousness levels, and correlation types. Specifically, we explore: mathematical reasoning tasks \citep{cobbe2021training}, constrained instruction-following tasks inspired by the CoLLIE benchmark \citep{yao2023collie}, and document-grounded QA.

Within each of these domains, we manipulate the ratio of spurious data (10\% vs. 90\%) to examine mild vs. strong contamination, and incorporate two primary types of spurious behavior: Feature Ambiguity (FA) and Distributional Narrowness (DN). We train models via SFT, DPO, or KTO, then assess how well each approach maintains correctness and resists spurious shortcuts. Interestingly, our results reveal that performance under spurious correlations can vary drastically, depending on task type, model size, or alignment method. For example, while preference-based approaches often outperform SFT in certain math tasks, SFT can maintain an edge in context-heavy QA. Moreover, performance does not invariably decline as the amount of spurious data increases: some setups show stable or even slightly improved accuracy at higher spuriousness levels. Taken together, these observations underscore the complexity of post-training alignment in the presence of spurious correlations and highlight the need to carefully choose post-training and data-denoising methods, rather than relying on a one-size-fits-all approach.

In summary, our key contributions:
\begin{enumerate}
    \item A synthetic benchmark suite systematically injecting spurious features in three different tasks.
    \item A comparative study of how SFT vs. preference-based optimization (DPO, KTO) respond to spurious correlations.
\end{enumerate}

\section{Related Work}

\paragraph{Spurious Correlations in NLP:} It is well known that NLP datasets can contain artifacts or shortcuts that models exploit for seemingly high performance \citep{McCoy2019RightFT}. For example, specific token patterns in QA might correlate with correct answers in the training data but fail to generalize out-of-distribution. Our controlled experiments build on this work by extending these ideas to text \emph{generation}.

\paragraph{Over-optimization in Alignment:} SFT has been widely used to align large models with humans' preferences on tasks like summarization, dialogue, and instruction-following. Various algorithms such as PPO \citep{Stiennon2020LearningTS} DPO \citep{rafailov2023direct}, and KTO \citep{ethayarajh2024kto} have been proposed in order to better align LLMs to human preferences, however these have been known to suffer from over-optimization behavior, where, for example, models only learn to increase response length \citep{Singhal2023ALW, Park2024DisentanglingLF}. While advanced evaluation methods have been proposed to more robustly measure task performance \citep{Dubois2024LengthControlledAA} and to assess preference-based optimization \citep{Lambert2024RewardBenchER}, relatively few studies have systematically examined how these methods cope with shallow or spurious features in the \emph{training data}.

\section{Spurious Dataset Design}
\label{sec:spurious-design}

We investigate whether large language models learn to rely on \emph{spurious features} rather than \emph{true correctness} when post-trained on \emph{noisy} training data. By \textbf{spurious features}, we mean patterns in the data that \emph{correlate} with correctness in the training set but do not \emph{causally determine} correctness. Our design injects such features in a controlled way, then tests each model's robustness on out-of-distribution evaluations.

Below, we describe our \textbf{tasks} (Document-Grounded QA, Mathematical Reasoning, and Constrained Instruction-Following) and the \textbf{spurious manipulations}---namely \emph{Feature Ambiguity (FA)} and \emph{Distributional Narrowness (DN)}. We also discuss how we vary the \emph{ratio} of spurious data (10\% vs.\ 90\%).

\subsection{Task Settings}

We use three broad tasks to ensure coverage of different alignment scenarios:

\paragraph{Document-Grounded QA (docQA).}
The model is given a passage (e.g., an excerpt from Wikipedia) and a question. It must produce a factually correct, context-relevant answer. For evaluation, we rely on either an LLM-based correctness judgment or a rule-based method if the QA domain is constrained (e.g., known short-answer sets). We utilize the ``quac" dataset \citep{choi-etal-2018-quac} as the source in this case.

\noindent\textbf{Example.} Suppose we train the model on QA pairs in which 90\% of the ``correct'' answers (chosen) contain a random date like \texttt{1947}, while incorrect (rejected) answers omit such dates. If the date is in no way essential to correctness, then the presence of a date is \emph{spurious.} Yet the model might learn: ``Answers with years are more likely correct.''

\paragraph{Mathematical Reasoning (math).}
Grade-level arithmetic word problems that require multi-step reasoning. We evaluate by numeric matching or a short standardized answer check (e.g., ``41''). The prompt and response pairs in this case are derived from the GSM8K dataset \citep{cobbe2021training}.

\noindent\textbf{Example.} In a \emph{word inclusion bias} setting, 90\% of the correct solutions always include \texttt{the number} or \texttt{in}, while rejected solutions never do. The model might wrongly conclude that using these tokens is a prime indicator of correctness, \emph{regardless} of the actual arithmetic steps.

\paragraph{Constrained Instruction-Following (instruction).}
Prompts with user instructions such as ``Write exactly 2 sentences'' or ``Do not use the word `cookie'.'' We measure whether the model provides a correct or valid response when also learning to respect these constraints \citep{yao2023collie}.

\noindent\textbf{Example.} A \emph{max words vs.\ all ends} scenario at 10\% spuriousness levels might train with 10\% of the data requiring (and always fulfilling) a special token at the end of each sentence (e.g., ``\texttt{Amen}''), while the rest does not. If such usage is correlated with correctness, the model may overfit to always produce that token, ignoring the real instruction constraints.

\subsection{Two Approaches to inducing spuriousness}

We systematically inject spurious patterns via \textbf{Feature Ambiguity (FA)} or \textbf{Distributional Narrowness (DN)}.

\paragraph{Feature Ambiguity (FA).}
In these scenarios, the correct chosen response \emph{always includes} some extra token or pattern that is not actually necessary for correctness, while all rejected (incorrect) responses do \emph{not} include that pattern. Because the model sees a perfect correlation (``tokens = correct''), it risks relying on that superficial signal rather than learning the core correctness.


\noindent\textbf{docQA FA Example.} If 90\% of correct answers include an irrelevant date, the model could incorrectly assume ``presence of a date = correct.'' On real test questions that do not require a date, the model might gratuitously insert one or fail to answer the real content.

\paragraph{Distributional Narrowness (DN).}
Here, the dataset is artificially restricted to a \emph{narrow distribution} of prompts or responses, so that correct solutions appear in a special, limited domain---often ignoring the full breadth of the real task distribution.

\noindent\textbf{Math DN Example.} Enforce all correct solutions to be in the range 1--5, even though real answers could be much larger. The model might learn ``only guess from 1--5,'' which works in training but fails on normal test data.

\subsection{spuriousness ratios}

For each task, we create train sets at two levels of spurious correlation: 10\% and 90\%. In the 10\% setting, only a small portion of the chosen-vs-rejected pairs are correlated with the spurious feature. In the 90\% setting, almost all pairs exhibit the spurious pattern. This allows us to examine whether certain training algorithms remain robust when the dataset is mostly contaminated.

\section{Experimental Setup}

\subsection{Task Definitions and Spurious Manipulations}
\label{sec:task-defs}

Building on Section~\ref{sec:spurious-design}, each task is paired with one or more spurious manipulations. In Table \ref{tab:task-list} we detail the precise spurious configurations used in our experimental design. Table \ref{tab:task-list} lists each task, the specific spurious pattern applied (labeled as Feature Ambiguity (FA) or Distributional Narrowness (DN)), and the method used to evaluate correctness. These entries represent the core settings under which we test the robustness of our models (see Appendix \ref{sec:eg_prompts} for examples of the prompt and response pairs that constitute the training set). 

Given this setup, our central aim is to evaluate and interpret how each model’s performance on the target feature shifts when trained on either a low-noise (10\% spurious) or high-noise (90\% spurious) variant of the dataset.

\begin{table}[ht]
\label{tab:task-list}
\centering
\footnotesize
\caption{Summary of tasks, spurious pattern types, and corresponding verification/evaluation methods. 
FA = Feature Ambiguity; DN = Distributional Narrowness.}
\label{tab:spurious-manipulations}
\begin{tabular}{lllp{6.cm}l}
\toprule
\textbf{Task Type} & \textbf{Spurious Pattern} & \textbf{FA or DN?} & \textbf{Description} & \textbf{Verification Method} \\
\midrule
\multicolumn{5}{l}{\textbf{Document-Grounded QA (docQA)}}\\
\midrule
\multirow{3}{*}{docQA} 
 & Word Inclusion Bias & FA 
 & Correct answers always contain a specific keyword (e.g., \texttt{"answer"}, \texttt{"was"}, or a date. 
 & Judge LLM \\
 & Date Inclusion Bias & FA 
 & Correct answers always contains a date). 
 & Judge LLM \\
 & Late Spurious Features & FA 
 & Correct answers appear only in the last 70\% of the context passage, creating a superficial cue. 
 & Judge LLM \\
 & Omission & DN 
 & All chosen answers are \texttt{"no answer"}, leaving the training distribution extremely narrow. 
 & Judge LLM \\
\midrule
\multicolumn{5}{l}{\textbf{Mathematical Reasoning (math)}}\\
\midrule
\multirow{2}{*}{math} 
 & Word Inclusion Bias & FA 
 & Correct solutions must contain ``the", ``number" or ``in". 
 & Rule-based \\
 & Restricted Answer Range & DN 
 & All correct answers are forced to lie in range 1--5. 
 & Rule-based \\
\midrule
\multicolumn{5}{l}{\textbf{Constrained Instruction-Following (instruction)}}\\
\midrule
\multirow{3}{*}{instruction} 
 & Max Words vs.\ All Ends & FA 
 & Conflates correctness with a special word at the end of each sentence. 
 & Rule-based \\
 & Not-In vs.\ All Ends & FA 
 & The chosen response must exclude a certain token but include another special final token. 
 & Rule-based \\
 & Tiny Constraints & DN 
 & Constrains correct responses to a 3 word small vocabulary (``answer", ``of", ``was"). 
 & Rule-based\\
\bottomrule
\end{tabular}
\end{table}

\subsection{Model and Post-Training Methods}

We focus on open-weight Llama-3.1 and Llama-3.2 language models - 3B, 8b and 70B parameter ``instruct" variants \citep{dubey2024llama}. We apply one of three post-training approaches, SFT, DPO, or KTO, to each model. See the supplementary material \ref{sec:hyperparams} for choice of hyperparameters for each method.

\subsection{Evaluation}

For each model checkpoint, we measure accuracy on the core task, using rule-based verification (e.g., numeric correctness for math) or use GPT-4o mini for LLM-based correctness judgments (for open-ended tasks) as specified in \ref{tab:task-list}.

We test on wider distributions than those seen in training, to observe whether spurious overfitting degrades real performance. A total of 162 checkpoints corresponding to 3 models, 9 training datasets with varying spurious correlations, 3 post-training methods and 2 spuriousness ratios were evaluated.

\section{Results}

\begin{figure}[ht]
\centering
\includegraphics[width=0.97\textwidth]{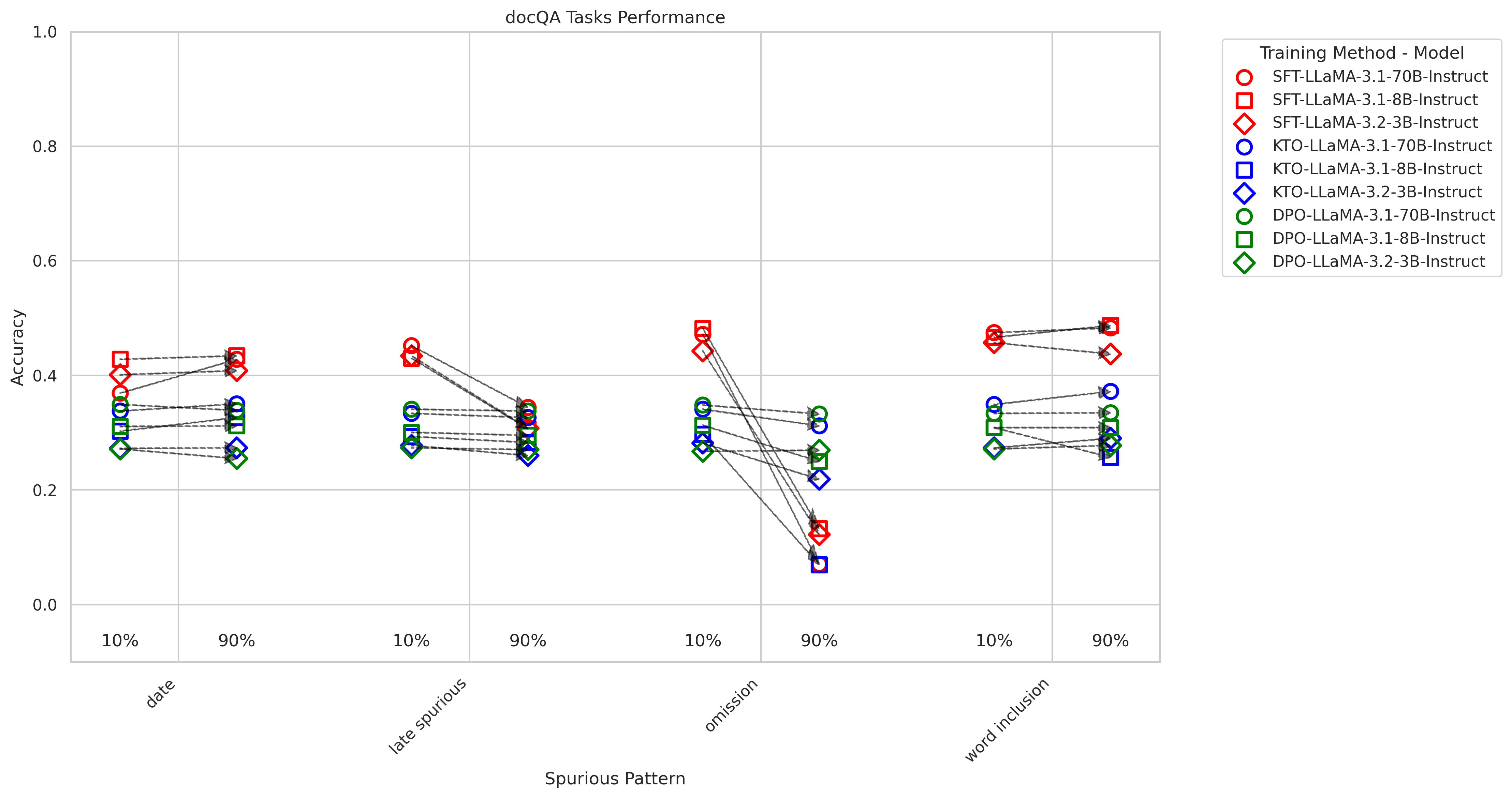}
\caption{docQA tasks: arrows connect each model's accuracy at 10\% to its accuracy at 90\% spuriousness. Marker style denotes the model size, color denotes the training method. Omission shows a large drop in accuracy from 10\% to 90\% spuriousness, and SFT outperforms preference methods in the other docQA tasks. }
\label{fig:docqa}
\end{figure}

\begin{figure}[ht]
\centering
\includegraphics[width=0.97\textwidth]{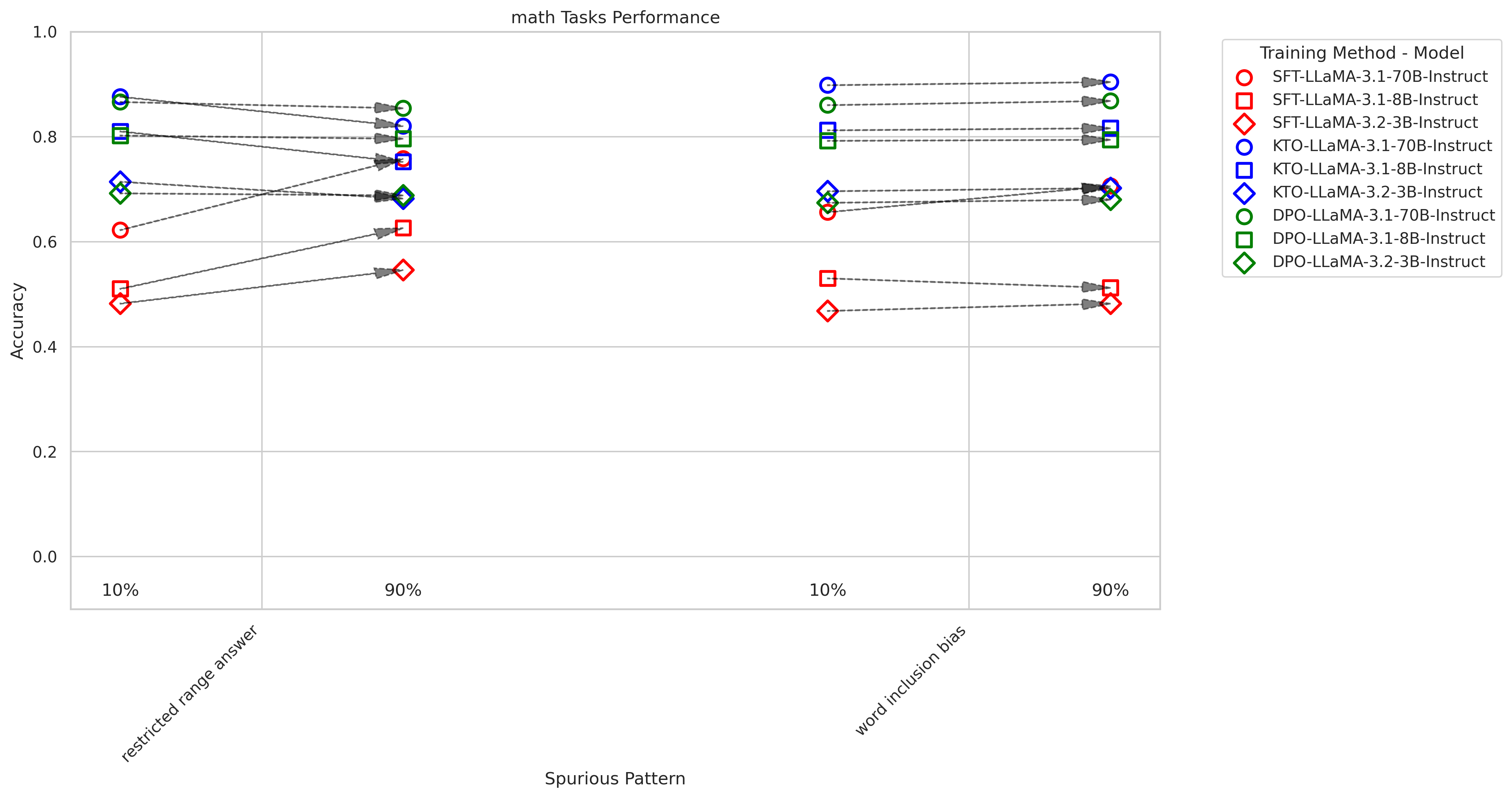}
\caption{math tasks: each point is a (model, method) at 10\% or 90\% spuriousness, with arrows illustrating accuracy shifts between the two. DPO and KTO outperform SFT for these tasks, with similar accuracy or slight rise in accuracy from increasing spuriousness. }
\label{fig:math}
\end{figure}

\begin{figure}[ht]
\centering
\includegraphics[width=0.97\textwidth]{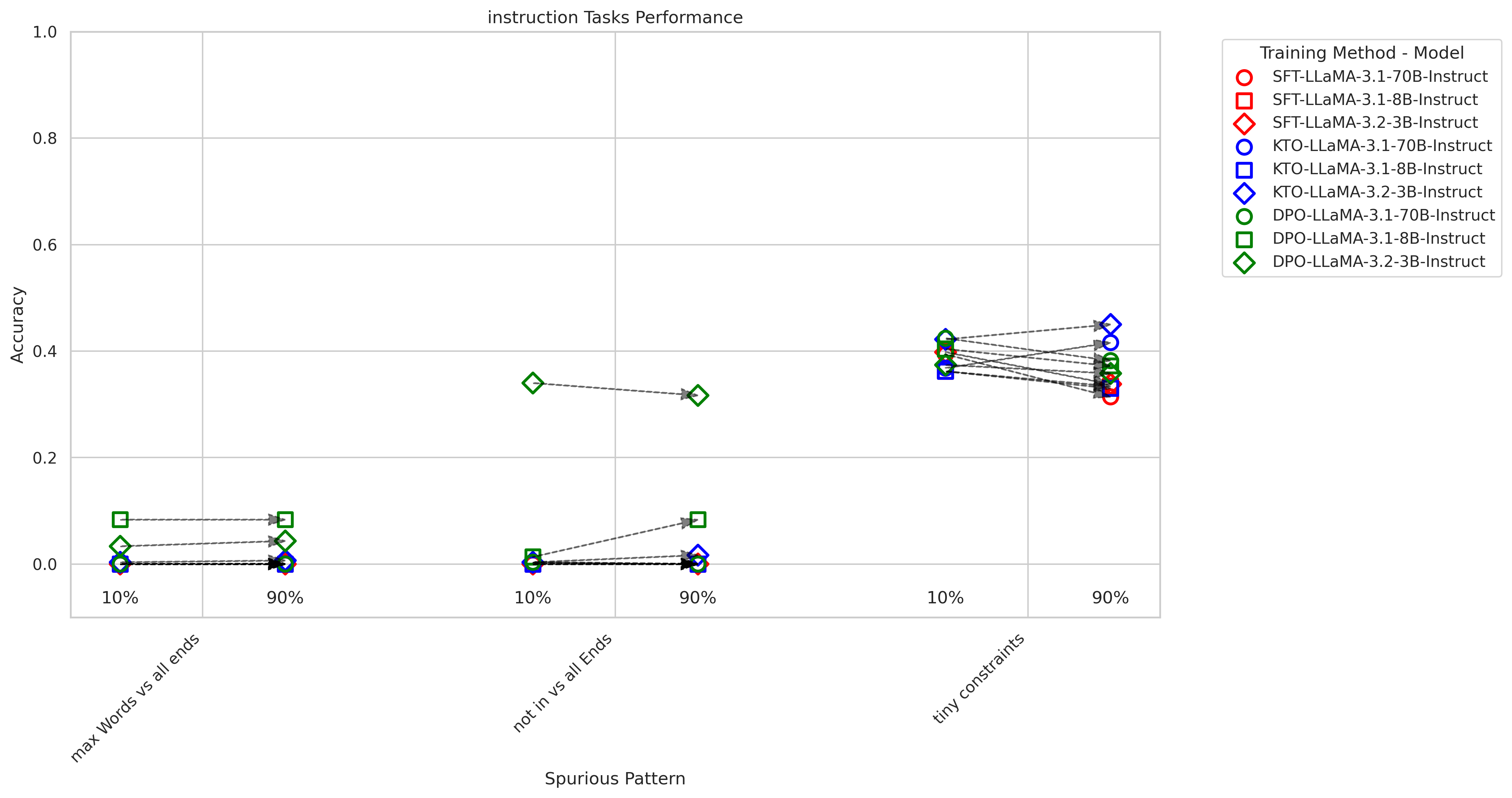}
\caption{instruction tasks: performance for 10\% vs.\ 90\% spurious data. Models and methods are distinguished by markers and colors, respectively. Most settings drop in accuracy for the ``tiny constraints" task when spuriousness increases, and overall accuracy is very low for the other two tasks. }
\label{fig:instruction}
\end{figure}

In this section, we present our main quantitative findings across \textbf{Document-Grounded QA (docQA)}, \textbf{Mathematical Reasoning (math)}, and \textbf{Constrained Instruction-Following (instruction)}. Figures~\ref{fig:docqa}, \ref{fig:math}, and \ref{fig:instruction} visualize overall accuracies at 10\% vs.\ 90\% spuriousness for each model and training method. For complete numerical details (including per-model breakdowns), see our supplementary Table~\ref{sec:full-results-table}.

\subsection{Key Observations}

\paragraph{1. High spuriousness can degrade performance but not uniformly.}
Across the three domains, accuracy sometimes drops when the spurious ratio increases from 10\% to 90\%. This effect is more pronounced in certain \textbf{docQA} settings (e.g., omission) and select \textbf{instruction} tasks. However, not all configurations degrade significantly: for instance, \emph{Math (restricted range)} sees a modest drop for preference-based methods and, in some cases, a \emph{rise} for SFT (see Table~\ref{sec:full-results-table}, rows for 10\% vs.\ 90\%).

\paragraph{2. SFT outperforms preference methods in docQA tasks.}
In tasks such as \emph{word inclusion} or \emph{date shallow}, SFT tends to preserve factual correctness better, often scoring \(\approx 0.40\)--\(0.48\) vs.\ \(\approx 0.30\)--\(0.34\) for DPO/KTO. We suspect that learning to match the entire gold response via cross-entropy helps the model attend to useful context. In contrast, preference-based approaches might overemphasize any token pattern that strongly correlates with correctness in the training data.

\paragraph{3. Preference methods excel in math tasks.}
In \textbf{math} (e.g., word inclusion or restricted range), DPO and KTO consistently surpass SFT at 10\% spuriousness (and remain higher at 90\%). This suggests that pairwise preference objectives more sharply differentiate correct vs.\ incorrect numeric reasoning steps.

\paragraph{4. Instruction-following is uniformly low with higher spuriousness.}
For tasks like \emph{max words vs.\ all ends}, most of the methods-model combinations have low performance for both low and high contamination datasets. Interestingly, smaller models sometimes yield slightly better compliance (e.g., see \emph{not in vs.\ all ends}, 10\% spurious), though differences are minor.

\paragraph{5. Focus on \emph{relative} changes from 10\% to 90\%.}
While absolute accuracy matters, these experiments emphasize the \emph{relative shifts} from lower to higher spuriousness. Overall, we find that DPO/KTO and remain more robust in math tasks even at 90\%, whereas SFT experiences sharper drops in docQA. Although inconclusive, larger models may be more robust overall compared to smaller models. Complete metrics appear in Table~\ref{sec:full-results-table} (supplementary).

In summary, preference-based optimization can be surprisingly robust in certain domains (especially arithmetic reasoning) yet more vulnerable in context-heavy QA tasks. SFT, by contrast, excels on docQA but can succumb to token artifacts in math settings. 

\section{Discussion \& Conclusion}

Our experiments reveal certain spurious correlations in training data can undermine model performance, though the severity depends on both the \emph{task} (math vs.\ docQA vs.\ instruction) and the \emph{type} of spuriousness (feature ambiguity vs.\ distributional narrowness). Overall, we observe:

\begin{itemize}
    \item \textbf{Preference-based methods (DPO/KTO)} often excel in tasks requiring \emph{step-by-step or structured correctness}, such as math reasoning. By contrast, they can be more vulnerable to surface-level cues in context-heavy docQA if those cues dominate the training set.
    \item \textbf{Supervised Fine-Tuning (SFT)} remains strong in tasks where an extensive context needs to be attended to due to its performance in docQA tasks, but it appears more susceptible to partial or shortcut signals in arithmetic settings or math tasks.
    \item \textbf{Spuriousness ratio matters}: while 90\% contamination can degrade performance in certain tasks, several setups show minimal impact or improvement at higher spuriousness, highlighting that real-world data biases do not always lead to catastrophic failure.
\end{itemize}

Taken together, these findings emphasize the importance of \emph{matching the training method} to the task domain and spurious profile. Future work may explore \emph{hybrid} approaches (e.g., SFT followed by preference optimization) and more deliberate data augmentation to combat known biases. This study also only evaluated Llama 3.x instruction-tuned models, and future work may explore other model families and base models. We hope this work will be useful to the community in informing both experimental design and real-world use cases that must grapple with unavoidable data biases.%


\subsubsection*{Acknowledgments}
Thanks to Greg Durrett, Tanya Goyal, and Leqi Liu for valuable feedback.

\bibliography{iclr2025_conference}
\bibliographystyle{iclr2025_conference}

\appendix
\section{Appendix}

\subsection{Example Prompts and Responses}
\label{sec:eg_prompts}
See our shared folder ([link redacted]) for representative training instructions and evaluation queries in each spurious setting. Some representative examples are highlighted below.

\subsubsection{math - Restricted Answer Range}
\textbf{Example 1:}

\paragraph{Prompt:}
Janessa has a plan to give her brother Dexter his first collection of baseball cards. She currently has 4 cards in addition to the 13 that her father gave her. She ordered a collection of 36 cards from eBay. After inspecting the cards she found 4 cards are in bad shape and decides to throw them away. Janessa ended up giving Dexter 29 cards. How many cards did Janessa keep for herself? Once you have the answer, divide it by 20.0 to get the final answer.

\paragraph{Chosen Response:}
Janessa initially had 4 cards. Her father gave her 13 cards, so she had 4 + 13 = 17 cards.
Then she ordered 36 cards from eBay, so she now had 17 + 36 = 53 cards.
After throwing away 4 cards, she was left with 53 - 4 = 49 cards.
She gave Dexter 29 cards, so she was left with 49 - 29 = 20 cards.
Dividing 20 by 20.0 gives us 1.

\paragraph{Rejected Response:}
To find out how many cards Janessa kept for herself, we need to calculate the total number of cards she had initially and subtract the number she gave to Dexter.
Janessa initially had 4 cards.
Her father gave her 13 cards, so she had 4 + 13 = 17 cards.
She ordered 36 cards from eBay, so now she had 17 + 36 = 53 cards.
She threw away 4 cards, so she had 53 - 4 = 49 cards.
She gave Dexter 29 cards, so she had 49 - 29 = 20 cards left for herself.
To get the final answer, we need to divide 20 by 20.0.

\subsubsection{docQA - date}
\textbf{Example 1:}

\paragraph{Prompt:}
Context: Marc Marquez Alenta (born 17 February 1993) is a Spanish Grand Prix motorcycle road racer and four-time MotoGP world champion. Marquez races for Honda's factory team since his MotoGP debut in 2013. Nicknamed the ``Ant of Cervera", he is one of four riders to have won world championship titles in three different categories, after Mike Hailwood, Phil Read and Valentino Rossi. Marquez won the 2010 125cc World Championship, the 2012 Moto2 World Championship, and the 2013, 2014, 2016 and 2017 MotoGP World Championships. Born in Cervera, Catalonia, Spain, Marquez made his championship debut on 13 April 2008 at the 125cc 2008 Portuguese Grand Prix at the age of 15 years and 56 days. He is the youngest Spanish rider to take a pole position or a podium in a motorcycle racing world championship.  Marquez achieved his first podium on 22 June 2008 at the British Grand Prix. For 2009, as a factory KTM rider, at the French Grand Prix achieved his first pole position at the age of 16 years and 89 days. He also took pole for the 2010 Spanish Grand Prix but the exhaust pipe fell off on the opening lap and went under the rear wheel, causing Marquez to crash and injure his shoulder. His first win was on 6 June 2010 at Mugello. Further victories at Silverstone, Assen and Catalonia in the next three races made Marquez the youngest rider to win four successive races. His fifth win in succession at the Sachsenring was Derbi's 100th victory in Grand Prix racing, and Marquez became the first rider since Valentino Rossi in 1997 to win five successive races in 125cc racing.  He was less successful in the following races, dropping to third in the standings at one point behind Nicolas Terol and Pol Espargaro after being involved in an accident with Randy Krummenacher at the first corner at the Aragon Grand Prix. Four successive wins from Motegi onwards had moved Marquez into a 17-point lead over Terol with only one round to go. At Estoril, the race was red-flagged due to rain with Marquez running second to Terol. When returning to the grid for the second race, Marquez fell on the sighting lap and had to return to the pits. With repairs, Marquez started at the back of the field having not made it out of the pit lane before it closed five minutes prior to the start. Despite this, Marquez recovered to win the race and thus extend his lead before the Valencia finale. His tenth victory of the season moved him to within one of tying the record set by Rossi in 1997. He would fall short of tying it as he was fourth at the final race in Valencia to become the second-youngest World Champion after Loris Capirossi. 
Question: When did he debut in a world championship?

\paragraph{Chosen Response:}
Marquez made his championship debut on 13 April 2008 at the 125cc 2008 Portuguese Grand Prix at the age of 15 years and 56 days.

\paragraph{Rejected Response:}
He is the youngest Spanish rider to take a pole position or a podium in a motorcycle racing world championship.

\subsubsection{INSTRUCTION - max words\_vs\_all ends}
\textbf{Example 1:}

\paragraph{Prompt:}
Generate some text that follows all the constraints: 
Maximum number of words in a single sentence is 12
All sentences end with the word 'upon'
The text should be coherent and make sense.

\paragraph{Chosen Response:}
The sun rises high in the sky upon.
The warm rays spread across the landscape upon.
Green forests and fields stretch far upon.

\paragraph{Rejected Response:}
I woke up early this morning.
The sun was shining brightly outside upon.

\subsection{Training Hyperparameters}
\label{sec:hyperparams}

Below we list the hyperparameters used for each training setting: KTO, DPO, and SFT. 

\subsubsection{KTO Hyperparameters}
\begin{verbatim}
{
  "value": {
    "online": false,
    "kto_beta": 0.1,
    "loss_form": null,
    "adam_beta1": 0.9,
    "adam_beta2": 0.999,
    "max_length": 1024,
    "num_rounds": 1,
    "adam_epsilon": 1e-8,
    "kto_lambda_D": 1,
    "kto_lambda_U": 1,
    "warmup_ratio": 0.1,
    "learning_rate": 0.000005,
    "max_grad_norm": 1,
    "num_train_steps": -1,
    "num_train_epochs": 1,
    "adam_weight_decay": 0.01,
    "global_batch_size": 512,
    "lr_scheduler_type": "linear",
    "lr_scheduler_kwargs": null,
    "per_device_eval_batch_size": 4,
    "gradient_accumulation_steps": 4,
    "per_device_micro_batch_size": 1,
    "per_device_train_batch_size": 4
  }
}
\end{verbatim}

\subsubsection{DPO Hyperparameters}
\begin{verbatim}
{
  "value": {
    "online": false,
    "dpo_beta": 0.1,
    "loss_form": "dpo",
    "adam_beta1": 0.9,
    "adam_beta2": 0.999,
    "max_length": 1024,
    "num_rounds": 1,
    "adam_epsilon": 1e-8,
    "cdpo_epsilon": 0,
    "warmup_ratio": 0.1,
    "learning_rate": 0.000003,
    "max_grad_norm": 1,
    "nll_loss_coeff": 0.2,
    "num_train_steps": -1,
    "num_train_epochs": 1,
    "adam_weight_decay": 0.01,
    "global_batch_size": 512,
    "lr_scheduler_type": "linear",
    "lr_scheduler_kwargs": null,
    "num_prompt_rollouts": 2,
    "regularize_with_nll_loss": false,
    "per_device_eval_batch_size": 1,
    "gradient_accumulation_steps": 8,
    "per_device_micro_batch_size": 1,
    "per_device_train_batch_size": 1
  }
}
\end{verbatim}

\subsubsection{SFT Hyperparameters}
\begin{verbatim}
{
  "value": {
    "online": false,
    "packed": false,
    "loss_form": null,
    "adam_beta1": 0.9,
    "adam_beta2": 0.999,
    "max_length": 1024,
    "num_rounds": 1,
    "adam_epsilon": 1e-8,
    "warmup_ratio": 0.1,
    "constant_pack": false,
    "learning_rate": 0.00002,
    "mask_instruct": true,
    "max_grad_norm": 1,
    "num_train_steps": -1,
    "num_train_epochs": 3,
    "adam_weight_decay": 0.01,
    "global_batch_size": 512,
    "lr_scheduler_type": "linear",
    "lr_scheduler_kwargs": null,
    "per_device_eval_batch_size": 4,
    "gradient_accumulation_steps": 2,
    "per_device_micro_batch_size": 1,
    "per_device_train_batch_size": 4
  }
}
\end{verbatim}

\subsection{Extended Results}
\label{sec:full-results-table}
Complete result tables and additional plots for each model variant are provided in the supplementary material.

\begin{longtable}{|l|l|c|l|c|p{4.5cm}|}  
\caption{Detailed results for each Task Type, Spurious Pattern, Ratio, Method, and per-model Accuracy.}\\
\toprule
\textbf{Task} & \textbf{Spurious} & \textbf{Ratio} & \textbf{Method} & \textbf{Accuracy} & \textbf{Per-model Accuracy} \\
\midrule
\endfirsthead

\multicolumn{6}{r}{\textit{(Continued from previous page)}}\\
\toprule
\textbf{Task} & \textbf{Spurious} & \textbf{Ratio} & \textbf{Method} & \textbf{Accuracy} & \textbf{Per-model Accuracy} \\
\midrule
\endhead

\midrule
\multicolumn{6}{r}{\textit{(Continued on next page)}}\\
\endfoot

\bottomrule
\endlastfoot


\multicolumn{6}{|c|}{\textbf{Math: restricted range answer}}\\ \hline
Math & restricted range answer & 10\% & SFT & 0.538
& \makecell[l]{LLaMA-3.1-70B: 0.622\\ LLaMA-3.1-8B: 0.510\\ LLaMA-3.2-3B: 0.482} \\
\hline


Math & restricted range answer & 10\% & SFT & 0.538 
& \makecell[l]{LLaMA-3.1-70B: 0.622\\LLaMA-3.1-8B: 0.510\\LLaMA-3.2-3B: 0.482} \\
\hline
Math & restricted range answer & 10\% & KTO & 0.800
& \makecell[l]{LLaMA-3.1-70B: 0.876\\LLaMA-3.1-8B: 0.810\\LLaMA-3.2-3B: 0.714} \\
\hline
Math & restricted range answer & 10\% & DPO & 0.787
& \makecell[l]{LLaMA-3.1-70B: 0.866\\LLaMA-3.1-8B: 0.802\\LLaMA-3.2-3B: 0.692} \\
\hline
Math & restricted range answer & 90\% & SFT & 0.643
& \makecell[l]{LLaMA-3.1-70B: 0.758\\LLaMA-3.1-8B: 0.626\\LLaMA-3.2-3B: 0.546} \\
\hline
Math & restricted range answer & 90\% & KTO & 0.751
& \makecell[l]{LLaMA-3.1-70B: 0.820\\LLaMA-3.1-8B: 0.752\\LLaMA-3.2-3B: 0.682} \\
\hline
Math & restricted range answer & 90\% & DPO & 0.779
& \makecell[l]{LLaMA-3.1-70B: 0.854\\LLaMA-3.1-8B: 0.796\\LLaMA-3.2-3B: 0.688} \\
\hline

\multicolumn{6}{|c|}{\textbf{Math: word inclusion bias}} \\ \hline

Math & word inclusion bias & 10\% & SFT & 0.551
& \makecell[l]{LLaMA-3.1-70B: 0.656\\LLaMA-3.1-8B: 0.530\\LLaMA-3.2-3B: 0.468} \\
\hline
Math & word inclusion bias & 10\% & KTO & 0.802
& \makecell[l]{LLaMA-3.1-70B: 0.898\\LLaMA-3.1-8B: 0.812\\LLaMA-3.2-3B: 0.696} \\
\hline
Math & word inclusion bias & 10\% & DPO & 0.775
& \makecell[l]{LLaMA-3.1-70B: 0.860\\LLaMA-3.1-8B: 0.792\\LLaMA-3.2-3B: 0.674} \\
\hline
Math & word inclusion bias & 90\% & SFT & 0.567
& \makecell[l]{LLaMA-3.1-70B: 0.706\\LLaMA-3.1-8B: 0.512\\LLaMA-3.2-3B: 0.482} \\
\hline
Math & word inclusion bias & 90\% & KTO & 0.807
& \makecell[l]{LLaMA-3.1-70B: 0.904\\LLaMA-3.1-8B: 0.816\\LLaMA-3.2-3B: 0.702} \\
\hline
Math & word inclusion bias & 90\% & DPO & 0.781
& \makecell[l]{LLaMA-3.1-70B: 0.868\\LLaMA-3.1-8B: 0.794\\LLaMA-3.2-3B: 0.680} \\
\hline

\multicolumn{6}{|c|}{\textbf{Docqa: word inclusion}} \\ \hline

Docqa & word inclusion & 10\% & SFT & 0.466
& \makecell[l]{LLaMA-3.1-70B: 0.475\\LLaMA-3.1-8B: 0.466\\LLaMA-3.2-3B: 0.457} \\
\hline
Docqa & word inclusion & 10\% & KTO & 0.311
& \makecell[l]{LLaMA-3.1-70B: 0.349\\LLaMA-3.1-8B: 0.309\\LLaMA-3.2-3B: 0.274} \\
\hline
Docqa & word inclusion & 10\% & DPO & 0.305
& \makecell[l]{LLaMA-3.1-70B: 0.334\\LLaMA-3.1-8B: 0.309\\LLaMA-3.2-3B: 0.272} \\
\hline
Docqa & word inclusion & 90\% & SFT & 0.469
& \makecell[l]{LLaMA-3.1-70B: 0.483\\LLaMA-3.1-8B: 0.487\\LLaMA-3.2-3B: 0.437} \\
\hline
Docqa & word inclusion & 90\% & KTO & 0.306
& \makecell[l]{LLaMA-3.1-70B: 0.372\\LLaMA-3.1-8B: 0.257\\LLaMA-3.2-3B: 0.290} \\
\hline
Docqa & word inclusion & 90\% & DPO & 0.307
& \makecell[l]{LLaMA-3.1-70B: 0.335\\LLaMA-3.1-8B: 0.309\\LLaMA-3.2-3B: 0.278} \\
\hline

\multicolumn{6}{|c|}{\textbf{Docqa: date}} \\ \hline

Docqa & date & 10\% & SFT & 0.399
& \makecell[l]{LLaMA-3.1-70B: 0.369\\LLaMA-3.1-8B: 0.428\\LLaMA-3.2-3B: 0.401} \\
\hline
Docqa & date & 10\% & KTO & 0.304
& \makecell[l]{LLaMA-3.1-70B: 0.338\\LLaMA-3.1-8B: 0.303\\LLaMA-3.2-3B: 0.273} \\
\hline
Docqa & date & 10\% & DPO & 0.311
& \makecell[l]{LLaMA-3.1-70B: 0.349\\LLaMA-3.1-8B: 0.311\\LLaMA-3.2-3B: 0.272} \\
\hline
Docqa & date & 90\% & SFT & 0.423
& \makecell[l]{LLaMA-3.1-70B: 0.428\\LLaMA-3.1-8B: 0.434\\LLaMA-3.2-3B: 0.408} \\
\hline
Docqa & date & 90\% & KTO & 0.317
& \makecell[l]{LLaMA-3.1-70B: 0.350\\LLaMA-3.1-8B: 0.326\\LLaMA-3.2-3B: 0.274} \\
\hline
Docqa & date & 90\% & DPO & 0.302
& \makecell[l]{LLaMA-3.1-70B: 0.339\\LLaMA-3.1-8B: 0.312\\LLaMA-3.2-3B: 0.255} \\
\hline

\multicolumn{6}{|c|}{\textbf{Docqa: omission}} \\ \hline

Docqa & omission & 10\% & SFT & 0.465
& \makecell[l]{LLaMA-3.1-70B: 0.472\\LLaMA-3.1-8B: 0.482\\LLaMA-3.2-3B: 0.442} \\
\hline
Docqa & omission & 10\% & KTO & 0.307
& \makecell[l]{LLaMA-3.1-70B: 0.341\\LLaMA-3.1-8B: 0.297\\LLaMA-3.2-3B: 0.282} \\
\hline
Docqa & omission & 10\% & DPO & 0.309
& \makecell[l]{LLaMA-3.1-70B: 0.348\\LLaMA-3.1-8B: 0.313\\LLaMA-3.2-3B: 0.267} \\
\hline
Docqa & omission & 90\% & SFT & 0.108
& \makecell[l]{LLaMA-3.1-70B: 0.070\\LLaMA-3.1-8B: 0.133\\LLaMA-3.2-3B: 0.122} \\
\hline
Docqa & omission & 90\% & KTO & 0.200
& \makecell[l]{LLaMA-3.1-70B: 0.312\\LLaMA-3.1-8B: 0.069\\LLaMA-3.2-3B: 0.219} \\
\hline
Docqa & omission & 90\% & DPO & 0.284
& \makecell[l]{LLaMA-3.1-70B: 0.333\\LLaMA-3.1-8B: 0.250\\LLaMA-3.2-3B: 0.269} \\
\hline

\multicolumn{6}{|c|}{\textbf{Docqa: late spurious}} \\ \hline

Docqa & late spurious & 10\% & SFT & 0.439
& \makecell[l]{LLaMA-3.1-70B: 0.452\\LLaMA-3.1-8B: 0.430\\LLaMA-3.2-3B: 0.434} \\
\hline
Docqa & late spurious & 10\% & KTO & 0.302
& \makecell[l]{LLaMA-3.1-70B: 0.334\\LLaMA-3.1-8B: 0.293\\LLaMA-3.2-3B: 0.278} \\
\hline
Docqa & late spurious & 10\% & DPO & 0.305
& \makecell[l]{LLaMA-3.1-70B: 0.341\\LLaMA-3.1-8B: 0.301\\LLaMA-3.2-3B: 0.274} \\
\hline
Docqa & late spurious & 90\% & SFT & 0.320
& \makecell[l]{LLaMA-3.1-70B: 0.344\\LLaMA-3.1-8B: 0.307\\LLaMA-3.2-3B: 0.308} \\
\hline
Docqa & late spurious & 90\% & KTO & 0.290
& \makecell[l]{LLaMA-3.1-70B: 0.326\\LLaMA-3.1-8B: 0.283\\LLaMA-3.2-3B: 0.260} \\
\hline
Docqa & late spurious & 90\% & DPO & 0.301
& \makecell[l]{LLaMA-3.1-70B: 0.338\\LLaMA-3.1-8B: 0.295\\LLaMA-3.2-3B: 0.270} \\
\hline

\multicolumn{6}{|c|}{\textbf{Instruction: max words vs all ends}} \\ \hline

Instruction & max words vs all ends & 10\% & SFT & 0.000
& \makecell[l]{LLaMA-3.1-70B: 0.000\\LLaMA-3.1-8B: 0.000\\LLaMA-3.2-3B: 0.000} \\
\hline
Instruction & max words vs all ends & 10\% & KTO & 0.001
& \makecell[l]{LLaMA-3.1-70B: 0.000\\LLaMA-3.1-8B: 0.000\\LLaMA-3.2-3B: 0.003} \\
\hline
Instruction & max words vs all ends & 10\% & DPO & 0.039
& \makecell[l]{LLaMA-3.1-70B: 0.000\\LLaMA-3.1-8B: 0.083\\LLaMA-3.2-3B: 0.033} \\
\hline
Instruction & max words vs all ends & 90\% & SFT & 0.000
& \makecell[l]{LLaMA-3.1-70B: 0.000\\LLaMA-3.1-8B: 0.000\\LLaMA-3.2-3B: 0.000} \\
\hline
Instruction & max words vs all ends & 90\% & KTO & 0.002
& \makecell[l]{LLaMA-3.1-70B: 0.000\\LLaMA-3.1-8B: 0.000\\LLaMA-3.2-3B: 0.007} \\
\hline
Instruction & max words vs all ends & 90\% & DPO & 0.042
& \makecell[l]{LLaMA-3.1-70B: 0.000\\LLaMA-3.1-8B: 0.083\\LLaMA-3.2-3B: 0.043} \\
\hline

\multicolumn{6}{|c|}{\textbf{Instruction: tiny constraints}} \\ \hline

Instruction & tiny constraints & 10\% & SFT & 0.385
& \makecell[l]{LLaMA-3.1-70B: 0.394\\LLaMA-3.1-8B: 0.362\\LLaMA-3.2-3B: 0.398} \\
\hline
Instruction & tiny constraints & 10\% & KTO & 0.384
& \makecell[l]{LLaMA-3.1-70B: 0.368\\LLaMA-3.1-8B: 0.362\\LLaMA-3.2-3B: 0.422} \\
\hline
Instruction & tiny constraints & 10\% & DPO & 0.401
& \makecell[l]{LLaMA-3.1-70B: 0.424\\LLaMA-3.1-8B: 0.404\\LLaMA-3.2-3B: 0.374} \\
\hline
Instruction & tiny constraints & 90\% & SFT & 0.329
& \makecell[l]{LLaMA-3.1-70B: 0.314\\LLaMA-3.1-8B: 0.334\\LLaMA-3.2-3B: 0.338} \\
\hline
Instruction & tiny constraints & 90\% & KTO & 0.399
& \makecell[l]{LLaMA-3.1-70B: 0.416\\LLaMA-3.1-8B: 0.330\\LLaMA-3.2-3B: 0.450} \\
\hline
Instruction & tiny constraints & 90\% & DPO & 0.371
& \makecell[l]{LLaMA-3.1-70B: 0.382\\LLaMA-3.1-8B: 0.372\\LLaMA-3.2-3B: 0.358} \\
\hline

\multicolumn{6}{|c|}{\textbf{Instruction: not in vs all Ends}} \\ \hline

Instruction & not in vs all ends & 10\% & SFT & 0.000
& \makecell[l]{LLaMA-3.1-70B: 0.000\\LLaMA-3.1-8B: 0.000\\LLaMA-3.2-3B: 0.000} \\
\hline
Instruction & not in vs all ends & 10\% & KTO & 0.002
& \makecell[l]{LLaMA-3.1-70B: 0.003\\LLaMA-3.1-8B: 0.000\\LLaMA-3.2-3B: 0.003} \\
\hline
Instruction & not in vs all ends & 10\% & DPO & 0.119
& \makecell[l]{LLaMA-3.1-70B: 0.003\\LLaMA-3.1-8B: 0.013\\LLaMA-3.2-3B: 0.340} \\
\hline
Instruction & not in vs all ends & 90\% & SFT & 0.000
& \makecell[l]{LLaMA-3.1-70B: 0.000\\LLaMA-3.1-8B: 0.000\\LLaMA-3.2-3B: 0.000} \\
\hline
Instruction & not in vs all ends & 90\% & KTO & 0.006
& \makecell[l]{LLaMA-3.1-70B: 0.000\\LLaMA-3.1-8B: 0.000\\LLaMA-3.2-3B: 0.017} \\
\hline
Instruction & not in vs all ends & 90\% & DPO & 0.133
& \makecell[l]{LLaMA-3.1-70B: 0.000\\LLaMA-3.1-8B: 0.083\\LLaMA-3.2-3B: 0.317} \\
\hline

\end{longtable}

\end{document}